\documentclass{article}

\usepackage{microtype}
\usepackage{graphicx}
\usepackage{subfig}
\usepackage{booktabs} 
\usepackage{amsmath,amsfonts,bm}
\usepackage{mathtools}
\usepackage{xcolor}
\usepackage{todonotes}

\DeclareMathOperator{\topk}{top-k}

\DeclarePairedDelimiter{\ceil}{\lceil}{\rceil}

\newcommand{\PLH}{{\mkern-2mu\times\mkern-2mu}}
\definecolor{blue(pigment)}{rgb}{0.2, 0.2, 0.6}
\definecolor{table_colour}{rgb}{0,0.08,0.45}

\usepackage{hyperref}

\usepackage[accepted]{icml2019}

\begin{document}

\twocolumn[
\icmltitle{On Graph Classification Networks, Datasets and Baselines }



\icmlsetsymbol{equal}{*}

\begin{icmlauthorlist}
\icmlauthor{Enxhell Luzhnica}{equal,to}
\icmlauthor{Ben Day}{equal,to}
\icmlauthor{Pietro Lio}{to}
\end{icmlauthorlist}

\icmlaffiliation{to}{Department of Computer Science \& Technology, University of Cambridge, Cambridge, United Kingdom}

\icmlcorrespondingauthor{Ben Day}{ben.day@cl.cam.ac.uk}

\icmlkeywords{Machine Learning, ICML}

\vskip 0.3in
]



\printAffiliationsAndNotice{\icmlEqualContribution} 

\begin{abstract}
    Graph classification receives a great deal of attention from the non-Euclidean machine learning community. Recent advances in graph coarsening have enabled the training of deeper networks and produced new state-of-the-art results in many benchmark tasks. We examine how these architectures train and find that performance is highly-sensitive to initialisation and depends strongly on jumping-knowledge structures. We then show that, despite the great complexity of these models, competitive performance is achieved by the simplest of models -- structure-blind \textsc{mlp}, single-layer \textsc{gcn} and fixed-weight \textsc{gcn} -- and propose these be included as baselines in future.
\end{abstract}

\section{Introduction}
Deep learning has produced remarkable results across the full breadth of machine learning research. For the most part this has been achieved through the reapplication of the two main architectures, the \textsc{cnn} and \textsc{rnn}, adapted to two Euclidean cases -- omnidirectional (image-like) and unidirectional (series) -- respectively. As such there is great interest in extending the general techniques to non-Euclidean cases and graph-structured data problems in particular.

These efforts are mostly inspired by the \textsc{cnn} and attempting to find suitable analogs to its core components, the convolutional and pooling operators. Early work set out to develop convolution-like graph operators. The focus has now turned to developing pooling operations, often referred to as coarsening in the context of graphs. Besides static methods \cite{Luzhnica2019CliqueClassification}, differentiable pooling frameworks have been developed. DiffPool achieved state-of-the-art (\textsc{s}o\textsc{ta}) performance across many benchmark tasks \cite{Ying2018HierarchicalPooling}, however a dense representation, quadratic in memory, is required. The Graph U-Net introduces a sparse method based on pruning nodes ($\topk$) \cite{gao2019graph}. Cangea et al. \yrcite{Cangea2018TowardsClassifiers} apply the method in graph classification by incorporating $\topk$ pools in a \textsc{gcn} model, achieving performance competitive with the \textsc{s}o\textsc{ta} with scalable memory requirements.

In this work we show that, under standard initialisation \cite{Glorot2010UnderstandingNetworks,He2015DelvingClassification}, using the \textsc{gcn} and $\topk$ operator together results in vanishing gradients beyond the first layers. In addition, we show that it is possible to attain good performance on smaller benchmark tasks simply using a global-pool\footnote{A simple mean or sum over the features of all nodes.} followed by an \textsc{mlp}. Furthermore, to achieve results on a par with Graph U-Net in \textit{all} benchmarks a single-layer \textsc{gcn} with a jumping-knowledge (\textsc{jk}) connection \cite{Xu2018RepresentationNetworks} from the input graph followed by an \textsc{mlp} is sufficient, whether the weights of the \textsc{gcn} are trained or not.

Considering the implications of these results, we primarily argue for the importance of including strong, simple baselines in evaluation. We also define an initialisation scheme that remedies the vanishing gradient issue by design though we find that this does not consistently improve performance.

\paragraph{Motivation}
This work was motivated by studies of network activations and gradient flow in deeper \textsc{gnn}s with \textsc{jk} structures and $\topk$ pooling. We found that, at initialisation, activations into the network rapidly vanish and that throughout training the gradients flowed mostly into earlier layers. These findings prompt two questions: firstly, are deeper networks only trainable thanks to \textsc{jk} structures bypassing later layers? and secondly, how important are the later layers to performance anyway?

\section{Preliminaries}
We use the standard notation: a graph $\mathcal{G}$ of $N$ nodes with $F$ features per node is represented by the pair $(\mathbf{A}, \mathbf{X})$ with adjacency matrix, $\mathbf{A} \in \mathbb{R}^{N \times N}$, and node feature matrix, $\mathbf{X} \in \mathbb{R}^{N \times F}$.

\paragraph{Graph Convolution} \textsc{R}e\textsc{lu} activations and the improved \textsc{gcn} \cite{gao2019graph} are used throughout. This differs from the standard \textsc{gcn} in that $\mathbf{\hat{A}}=\mathbf{A}+2\mathbf{I}$ is used i.e. self-loops have a weight of 2.

\paragraph{Pooling} $\topk$ pooling is used \cite{gao2019graph}. The pooling operator drops $N - \ceil{kN}$ nodes, where $k \in [0,1)$ is a fixed hyperparameter. In all experiments this was set to $0.8$. Nodes are dropped based on the ranked projection of features on a learnable vector, $\Vec{p}$, as
\begin{align}
    \hat{y}_i &= \frac{\textup{X}_i\cdot\Vec{p}} {\| \Vec{p} \|} &\Vec{i} = \topk(\Vec{y}, k) \nonumber \\
    \mathbf{X'} &= \textup{X}_{\Vec{i}} \odot \tanh(\Vec{y}_{\Vec{i}}) &\mathbf{A'} = \mathbf{A}_{\Vec{i};\Vec{i}} \nonumber
\end{align}
where $\hat{y}$ are the scores for each node (rows in $\mathbf{X}$) and $\Vec{i}$ are the indices of the top-$k$ nodes based on their scores.

\paragraph{Jumping Knowledge Networks}
In node aggregating schemes, the range of nodes\footnote{Analogous to the receptive field in \textsc{cnn}s.} that a node's representation draws from is strongly dependent on the neighbourhood structure \cite{Xu2018RepresentationNetworks}. \textsc{Jk}-structures were introduced to allow some flexibility over the degree of aggregation and thus even out the ``range'' by introducing layer skipping connections. For a node, $v$, this takes the form
\begin{align}
    h_v^{1} &= f_1(X_v) \quad ; \quad h_v^{i} = f_i(h_v^{i-1})\nonumber \\
    h_v^{JK} &= \textup{Agg.}(h_v^1,\dots,h_v^L) \nonumber
\end{align}
where the aggregation function is typically concatenation, summation or an elementwise max, the result being passed to a classifier.


\section{Removing JK \& Initialisation}
\label{remove-jk}
Whilst \textsc{jk}-connections were introduced to tackle the problem of node-specific range, in deeper networks they are acting as bypasses of later layers and a hierarchy of representations is not actually being produced. Clearly it runs counter to the core concept of allowing the range to vary over nodes if the higher ranges are not used. To test this we expose the gradient flow and activations in a net of four blocks of \textsc{gcn}+$\topk $ with the final representation aggregated with a global mean and entered into an \textsc{mlp}. Re\textsc{lu} activations are used in the \textsc{gcn}. The \textsc{gcn} weights are initialised using Kaiming \cite{He2015DelvingClassification}, while the pools are initialized using Glorot \cite{Glorot2010UnderstandingNetworks}\footnote{The authors note the mixed naming conventions here but this seems to be what the community has settled on.}. We refer to this combination as the `standard initialisation'. Under standard initialisation, layer activations decay into the network, gradients are vanishingly small and the latter part of the network is effectively static under backpropagation.

\subsection{\textsc{ReInit}}
To expose this problem we propose a data-driven approach similar to \textsc{lsuv}-initialisation \cite{Mishkin2015AllInit} to maintain variance across layers. The idea is simply to initialise under some scheme and then pass the entire batch through each block, scaling the layer weights in turn by $\sigma^{-1}$ to maintain variance, a process we refer to as \textsc{ReInit}. This is implemented as scaling factors that are set progressively
\begin{align}
    \mathbf{X'} &= \frac{1}{c_1} \textup{GCN}(\mathbf{X},\mathbf{A}) \quad ; \quad c_1 = \sigma\big(\textup{GCN}(\mathbf{X},\mathbf{A})\big) \nonumber \\
    \mathbf{X''} &= \frac{1}{c_2} \textup{X}'_{\Vec{i}} \odot \tanh(\Vec{y}_i) \quad ; \quad c_2 = \sigma\big(\textup{X}'_{\Vec{i}} \odot \tanh(\Vec{y}_i)\big) \nonumber
\end{align}
with the result that $\sigma(\mathbf{X}')=\sigma(\mathbf{X}'')=1$. We deviate from \textsc{lsuv} in not ortho-normalising as there is not an analogue that could be applied to the $\topk$ layers so simply rescaling has a more consistent meaning over the network. We have also found that attempting to derive a semi-analytic solution, in the footsteps of Glorot \& Bengio \yrcite{Glorot2010UnderstandingNetworks}, is not possible for the \textsc{gcn} due to the structural asymmetries in neighbourhood aggregation. In essence, the expected variance is sensitive to the number and similarity of neighbours to such a degree that properly accounting for these variations would require specific node-level information. This also allows \textsc{ReInit} to be applied on top of any initial scheme, so the `shape' is not fixed in that sense.

\section{Shallower, Simpler Networks}
\label{shallow-simple}
To see how much later \textsc{gcn} layers contribute to performance we tested three shallower networks on standard benchmarks. The models could be thought of as extreme ablations.

\paragraph{Structure-blind MLP}
A three-layer \textsc{mlp}. The adjacency matrix is discarded, the features are globally pooled and passed as input. Three weight matrices, biases; \textsc{r}e\textsc{lu} activations. This model cannot see even the number of nodes let alone their individual features or structural relationships.

\paragraph{Single-layer JK GCN+MLP}
A single layer \textsc{gcn} with a \textsc{jk}-skip preceding the \textsc{mlp} described above. We test this set up both with the weights of the \textsc{gcn} fixed at the random initialization values, denoted \textsc{(r)}, and free to update. The fixed method is intended to provide a minimal structural addition to the plain \textsc{mlp}.

\section{Experiments \& Results}

\begin{figure}
\begin{center}
\includegraphics[width=\columnwidth]{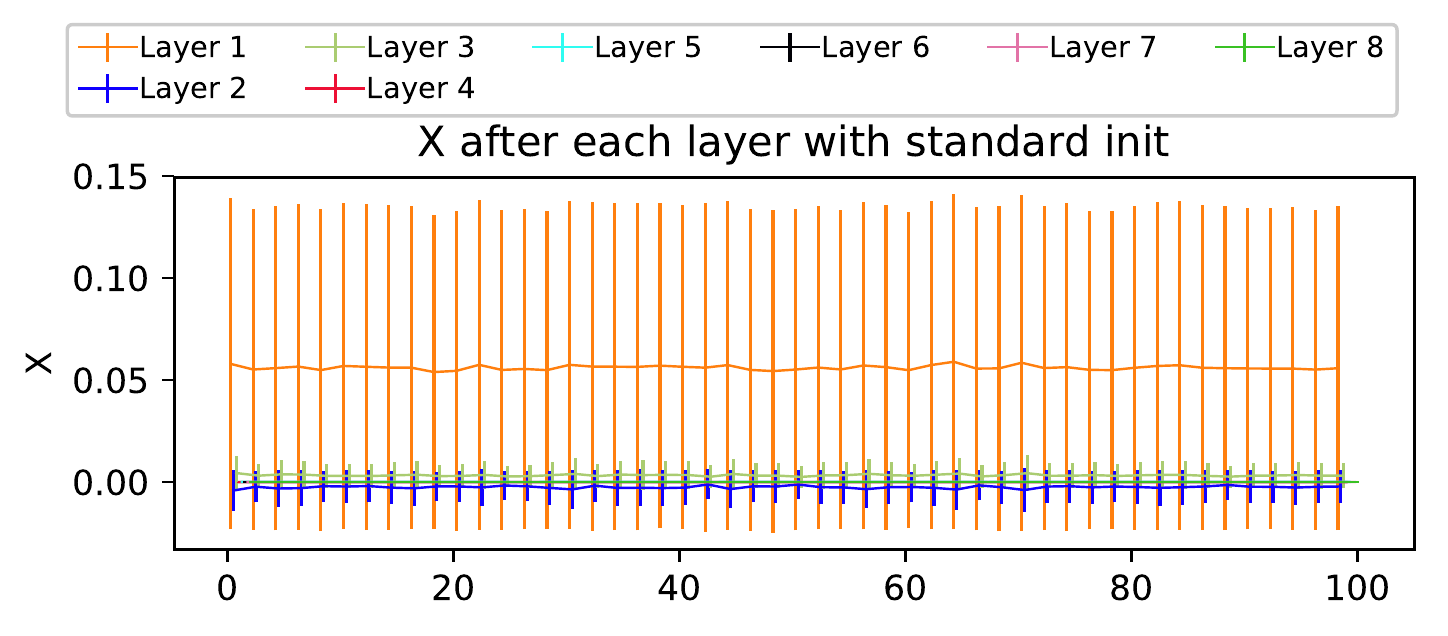}
\includegraphics[width=\columnwidth]{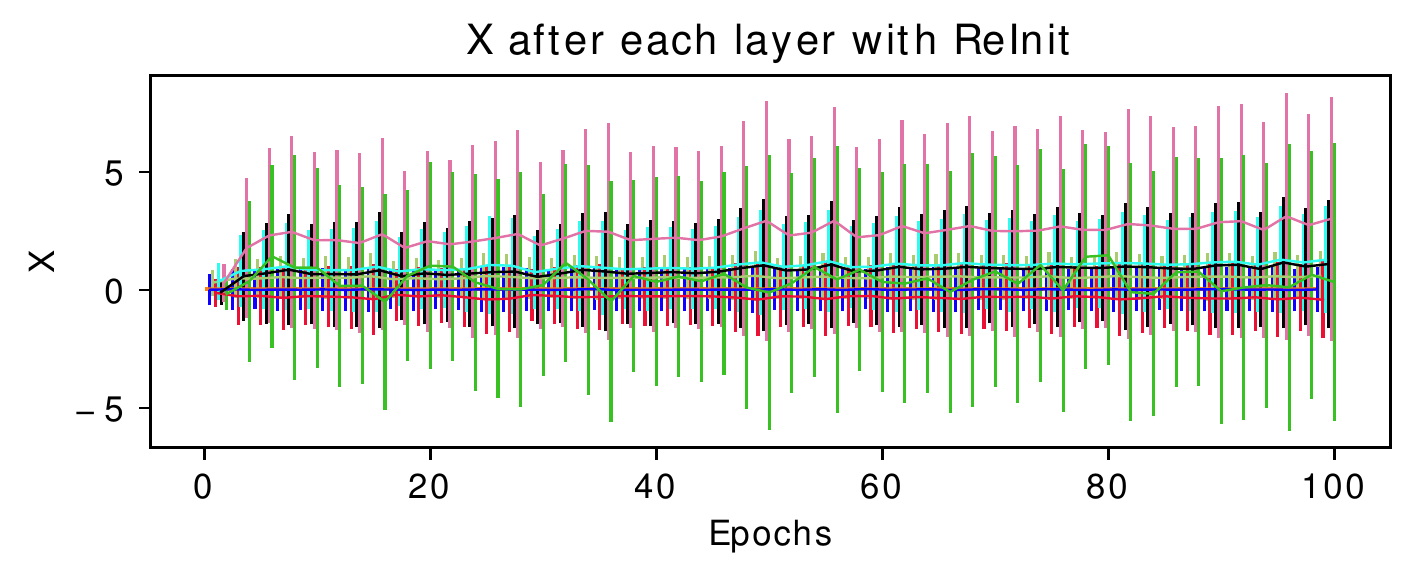}
\caption{Outputs of each layer during training with the standard initialization (top) and ours (bottom). Note the scale difference.
The standard initialization quickly converges to zero for all layers, while with \textsc{ReInit} the values vary widely }
\label{linear-activations}
\end{center}
\vskip -0.5in
\end{figure}

\begin{figure}
\begin{center}
\includegraphics[width=\columnwidth]{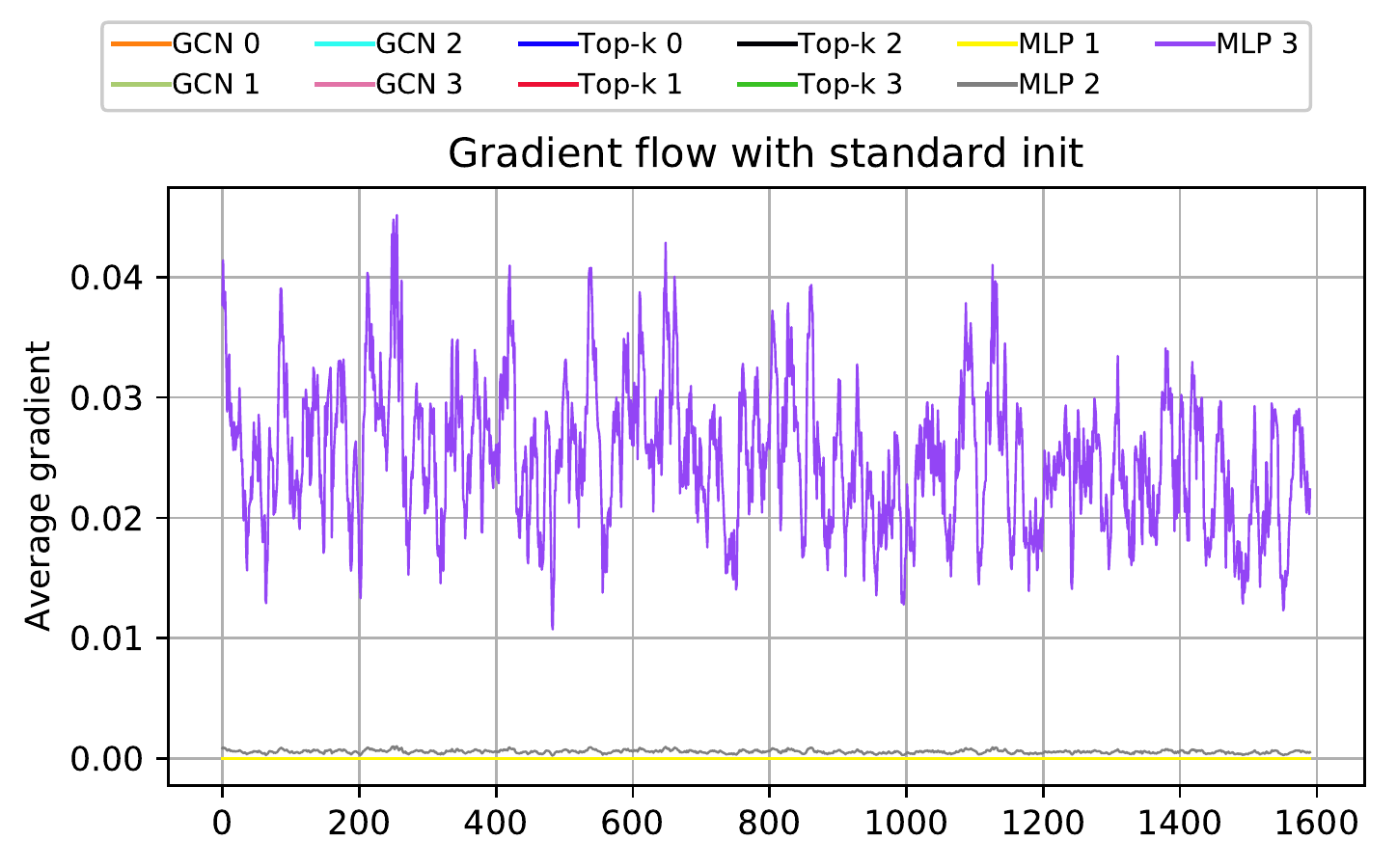}
\includegraphics[width=\columnwidth]{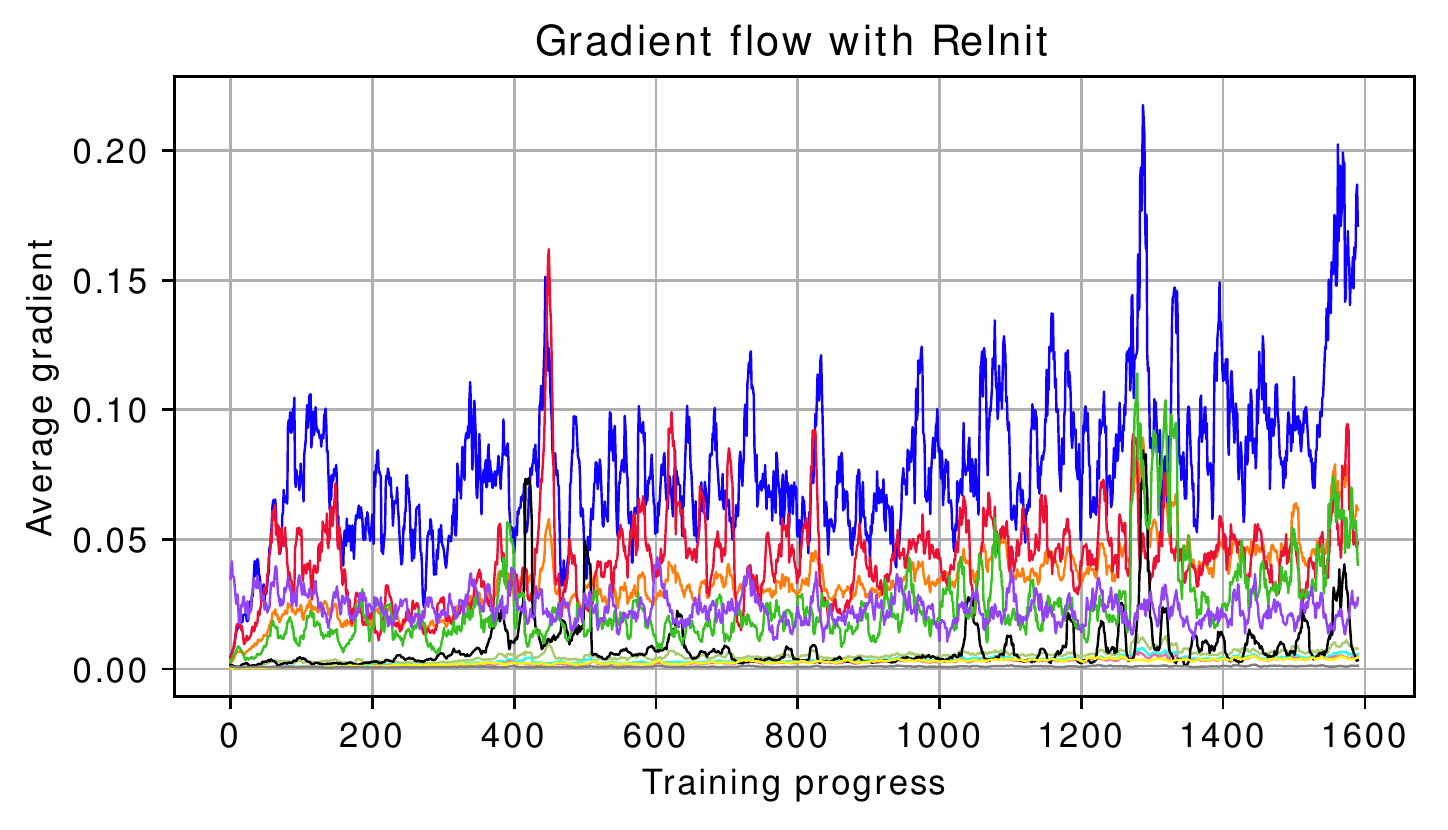}
\caption{Gradients flowing into the weights of all layers with regular initialization (top) and \textsc{ReInit} (bottom). The gradients of all the layers apart from the last MLP layer are almost 0 for the regular initialization. The reinitialized network manages to train the other layers, although noticeably less gradients flow into the latter layers, possibly by choice rather than a network problem.}
\label{linear-gradients}
\end{center}
\vskip -0.4in
\end{figure}

\begin{figure}
\begin{center}
\centerline{\includegraphics[scale=0.3]{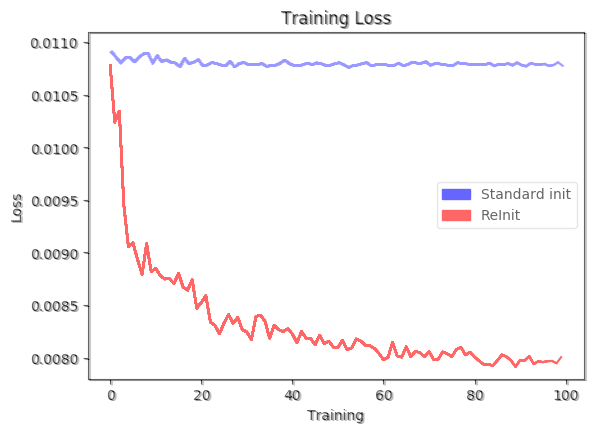}}
\caption{Training loss for the standard initialization and \textsc{ReInit}. The loss does not change for the standard initialization while with \textsc{ReInit} the network is successfully trained.}
\label{linear-training-loss}
\end{center}
\vskip -0.4in
\end{figure}

\begin{figure}[ht!]
\begin{center}
\centerline{\includegraphics[width=\columnwidth]{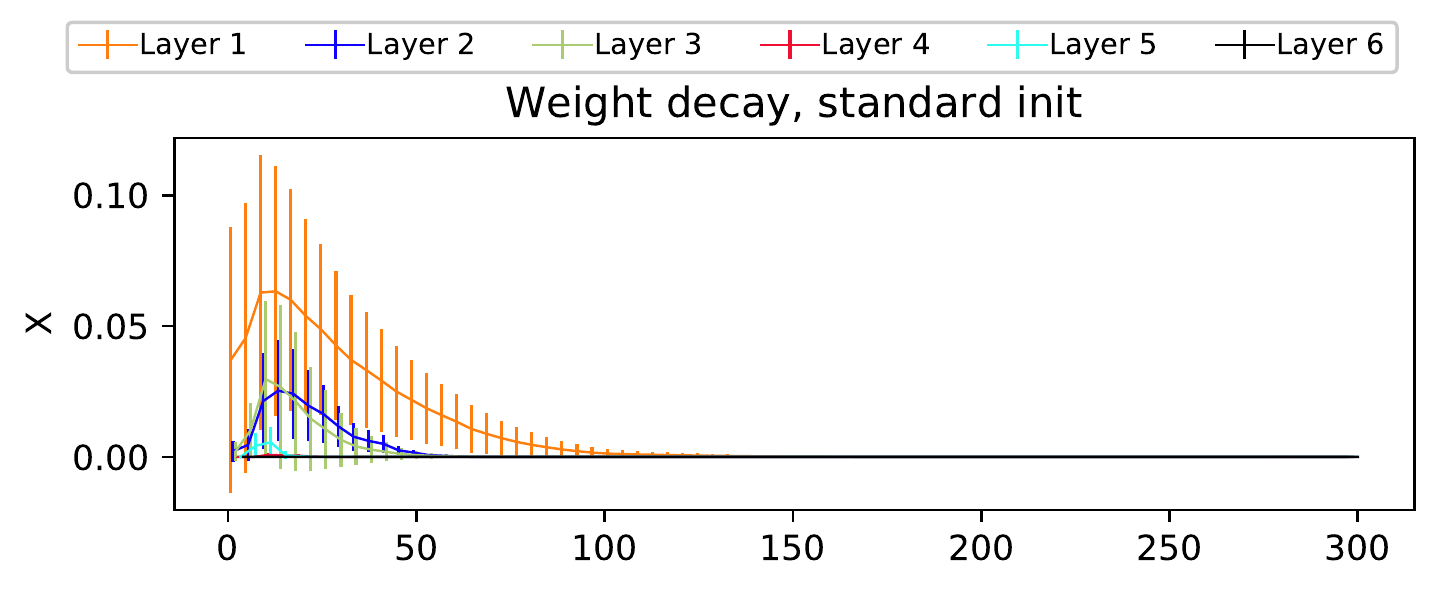}}
\includegraphics[width=\columnwidth]{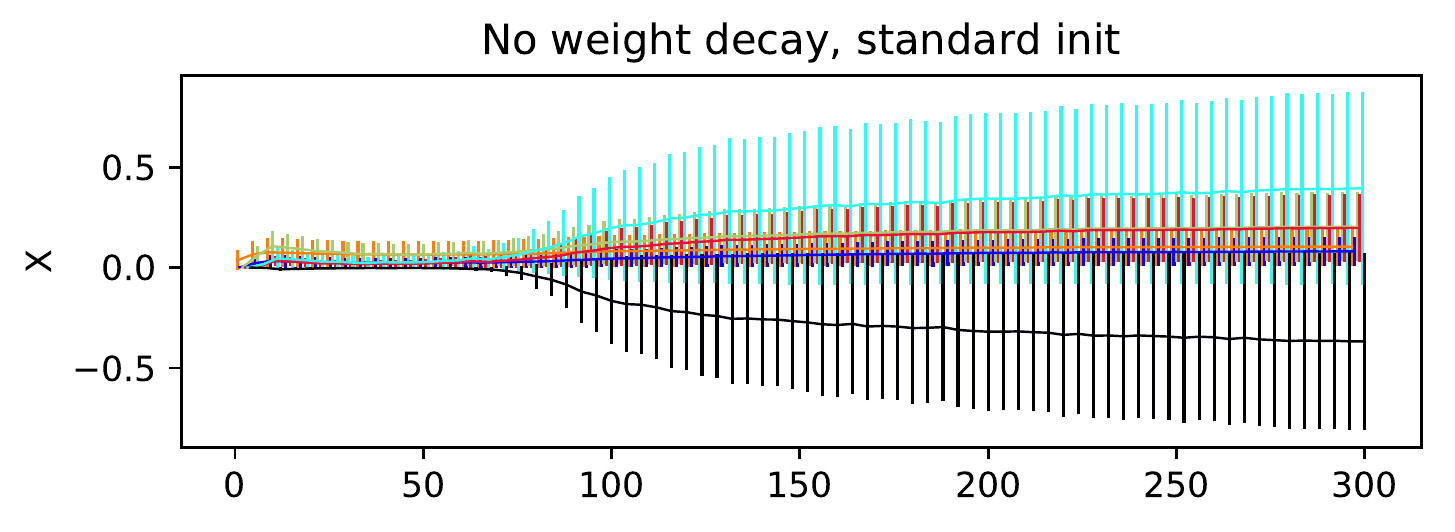}
\centerline{\includegraphics[width=\columnwidth]{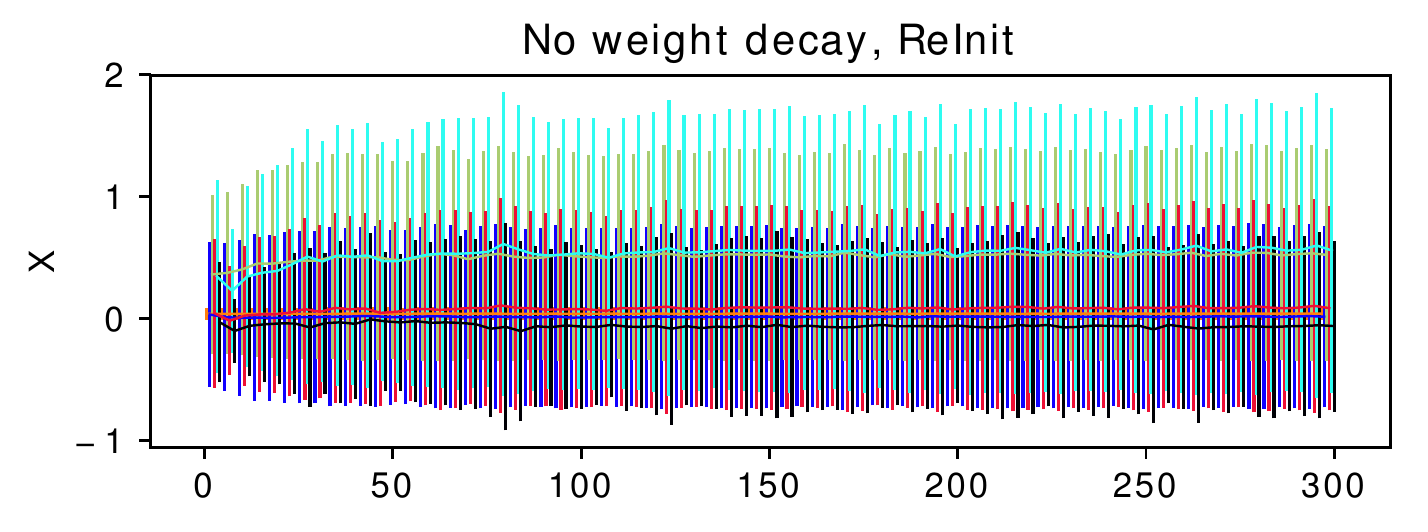}}
\centerline{\includegraphics[scale=0.5]{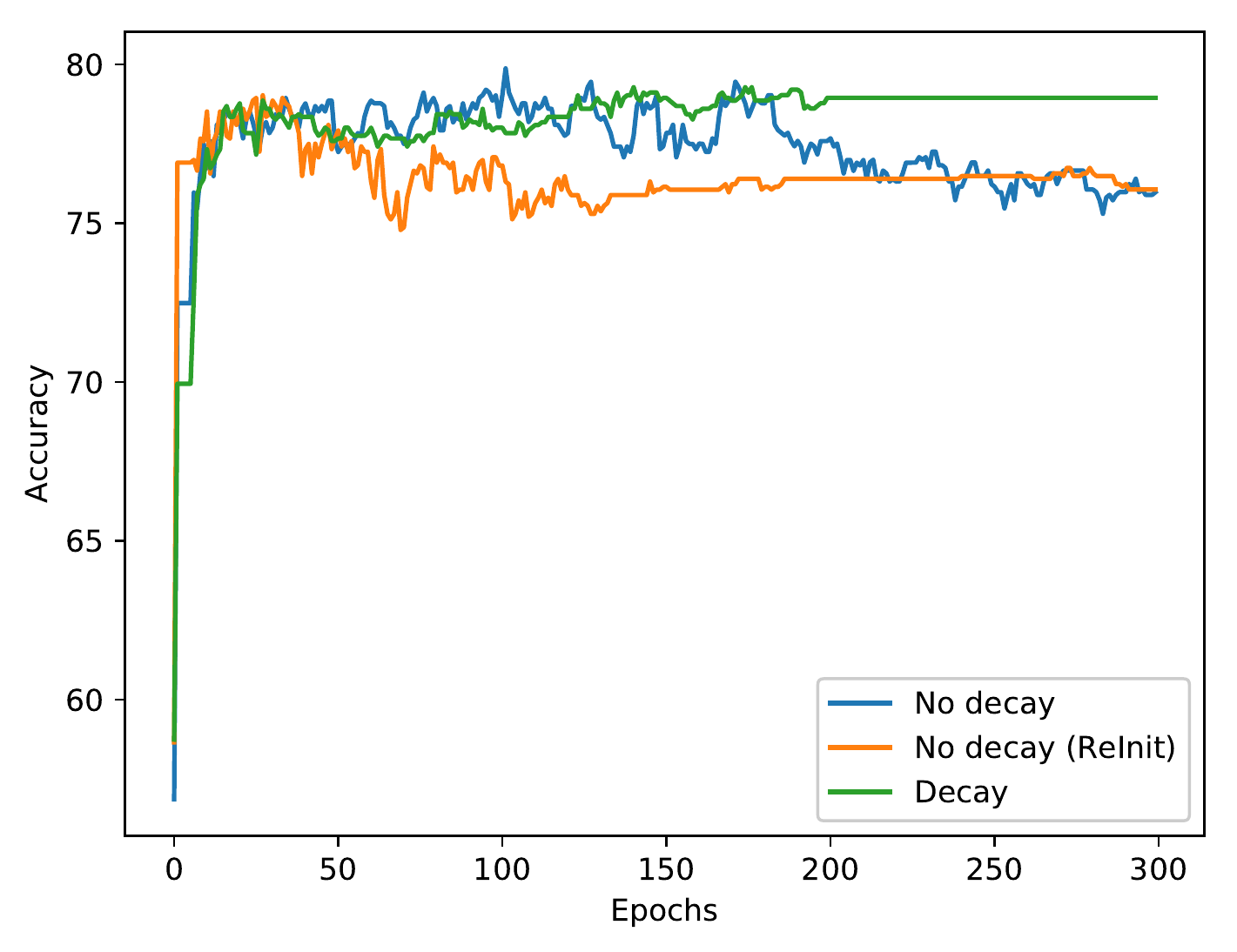}}

\caption{Output values in different training and initialisation routines when training for 300 epochs on the \textsc{dd} dataset. The first plot shows  pre-activations vanish in a simple \textsc{jk}-net under standard initialization, trained with Adam with weight decay. The second shows the same network trained without weight decay. The third has no weight decay and is initialized with \textsc{ReInit}. The last figure shows the performance of the three setups on the \textsc{dd} dataset (over 10 folds) as we vary the number of epochs.}
\label{activations-jk}
\end{center}
\vskip -0.2in
\end{figure}

\paragraph{No JK}
We first present the comparison of activations, gradient flow and training dynamics for a 4-block \textsc{gnn} (as described in \ref{remove-jk})  in figures \ref{linear-activations}, \ref{linear-gradients} \& \ref{linear-training-loss}, respectively. Detailed analysis of these plots is presented as captions, though the overall picture is that under \textsc{ReInit} training is able to occur whilst under standard initialisation it is not.

\subsection{Shallow baselines}
We conduct several experiments with the networks described in section \ref{shallow-simple}: a simple \textsc{mlp}; a randomly initialized \textsc{gcn}, which is not updated during the training process, denoted \textsc{gcn(r)-mlp}; and a \textsc{gcn} that is free to update (\textsc{gcn-mlp}).

We find that these models surpass most of the previous methods. In some cases surpassing even the recent differentiable pooling methods. We note that the performance of the random \textsc{gcn} should not come as a surprise given its connection to WL-test \cite{Kipf2016Semi-SupervisedNetworks}. This is most relevant in the case of the random \textsc{gcn}, having very little power in the featural domain but adding structural information comparable to 1-WL.

These initial results (presented in table \ref{table:results}) show that there is room for advancements in graph classification and that these simple models are to be considered strong baselines. These networks, particularly the \textsc{mlp}, are simple and appear as subnetworks in many methods. As such, it is of paramount importance to undertake thorough ablation studies to show the benefit of complexifying networks. For instance, we can add additional components that improve upon other approaches but do so by relying heavily on these simpler subnetworks. We explore this idea below.

\subsection{Bloated networks}
We use the following architecture in the next few experiments: \textsc{gcn-pool-gcn-pool-gcn-pool-mlp} with the global max and sum of each layer passed to the \textsc{mlp} through \textsc{jk}-structures. Due to the initialization problem, if weight decay is used\footnote{Here we use $\lambda = 5\PLH10^{-3}$ with a learning rate of $5\PLH10^{-4}$ but smaller values achieve similar results.} the network is unable to recover from a bad initialization and as such it cannot learn in the deeper layers (see Figure \ref{activations-jk}). This method (\textsc{jk-sum-decay}) is competitive with most results, performing closely to the simple sub-network it contains: \textsc{gcn-mlp}.

\begin{table}
\vskip 0.15in
\centering
\caption{Classification accuracy percentages. The results of other networks are taken from \citealt{Cangea2018TowardsClassifiers} with which we share 10-fold splits for benchmarking our methods. Bold indicates top-performance, \textcolor{table_colour}{blue} indicates weaker performance than the \textsc{mlp}.}
\begin{small}
\begin{sc}
\begin{tabular}{lcccr}
\toprule
 & \multicolumn{4}{c}{Datasets} \\
 \cmidrule{2-5}
Model & Reddit \footnote{Reddit-Multi-12K} & DD &  Collab & Prot.\\
\midrule
PatchySAN       & 41.32         & \textcolor{table_colour}{76.27}& \textcolor{table_colour}{72.60} & \textcolor{table_colour}{75.00}\\
GraphSAGE       & 42.24         & \textcolor{table_colour}{75.42}& \textcolor{table_colour}{68.25} & \textcolor{table_colour}{70.48}\\
ECC             & 41.73         & \textcolor{table_colour}{74.10}& \textcolor{table_colour}{67.79} & \textcolor{table_colour}{72.65}\\
Set2Set         & 43.49         & \textcolor{table_colour}{78.12}& \textcolor{table_colour}{71.75} & \textcolor{table_colour}{74.29} \\
SortPool        & 41.82         & \textcolor{table_colour}{79.37}& \textcolor{table_colour}{73.76} & \textcolor{table_colour}{75.54}\\
DiffPool-Det    & 46.18         & \textcolor{table_colour}{75.47}& \textbf{82.13} & \textcolor{table_colour}{75.62}\\
DiffPool-NoLP   &  46.65        & \textcolor{table_colour}{79.98}&  75.63 & 77.42\\
DiffPool        & 47.08         & \textbf{81.15}& 75.50 & \textbf{78.10}\\
GU-Net/SHGC     & -             & \textcolor{table_colour}{78.59}& 74.54 & \textcolor{table_colour}{75.46}\\
\midrule
MLP          & 40.96         & 80.22         & 74.00 & 75.74\\
GCN(R)-MLP    & \textcolor{table_colour}{36.15}& \textcolor{table_colour}{78.61}& 75.38 & 76.28\\
GCN-MLP       & 45.01         & \textcolor{table_colour}{79.29}& 76.50 &  \textcolor{table_colour}{75.64}\\
\midrule
JK-Sum          & \textbf{47.16}         & \textcolor{table_colour}{79.02}& 77.00 & 75.82\\ 
JK-Sum-Decay    & 43.87         & \textcolor{table_colour}{79.11}& 74.14 & 75.82\\
JK-Sum-ReInit   & 46.77& \textcolor{table_colour}{75.97}& 77.20 & \textcolor{table_colour}{75.46}\\
\bottomrule
\end{tabular}
\end{sc}
\end{small}
\label{table:results}
\end{table}

Next, even if we do not use any weight decay the network will only be able to recover the deeper layers after a significant number of epochs. For instance, for DD the network only starts to recover the deeper layers after epoch $100$ as shown in Figure \ref{activations-jk}. Although, to fully recover the layers (similarly to the network with \textsc{ReInit}) we found that the network needs to be trained for more than $800$ epochs and, if early-stopping causes training to end in an earlier epoch, we would still be using only the first two layers (\textsc{gcn}+\textsc{pool}). In fact, the optimal number of epochs to train the network for was $100$ which is what we report in the results in Table \ref{table:results} (\textsc{jk-sum}). However, the network behaves very differently when initialized using \textsc{ReInit} as the method does not need to recover the layers one-by-one, changing the dynamics and ultimately how and what the network learns. The same figure shows that in the case of \textsc{ReInit} all the layers are trainable from the beginning. In that case, we notice that the performance goes up sharply in the very first few epochs for DD (less than 10, see last plot of Figure \ref{activations-jk}) and then drops and converges to roughly the same as the recovered network with standard initialization (without weight decay). While for small datasets (\textsc{DD, Proteins}) unleashing the power of the deeper network from the beginning is not beneficial since it can cause over-fitting (a single layer \textsc{gcn} already performs well) for \textsc{collab} we see that this differs. In fact, for these small  datasets, the method with \textsc{ReInit} achieved highest accuracy in fewer than 50 epochs, while for \textsc{Collab} it was 300. The same network without \textsc{ReInit} had the best performance training for 100 epochs, but resulted in a lower quality model. This hints that for this bigger dataset all 3-layers are needed, while for smaller problems the network is likely over-parameterised and this is exposed by \textsc{ReInit}.

\paragraph{Closing remarks} We have demonstrated that some very simple models are competitive with the \textsc{s}o\textsc{ta} and that \textsc{jk}-structures may permit models to perform well through these subnetworks. We hope that these baselines and a greater interest in ablation studies will be adopted by the community.

\clearpage
\nocite{Bronstein2017GeometricData}
\nocite{pmlr-v15-glorot11a,zhang2018end,gilmer2017neural,niepert2016learning,Simonovsky_2017,hamilton2017inductive,He_2016,Fey/Lenssen/2019}

\bibliography{example_paper,references}

\begin{thebibliography}{18}
\providecommand{\natexlab}[1]{#1}
\providecommand{\url}[1]{\texttt{#1}}
\expandafter\ifx\csname urlstyle\endcsname\relax
  \providecommand{\doi}[1]{doi: #1}\else
  \providecommand{\doi}{doi: \begingroup \urlstyle{rm}\Url}\fi

\bibitem[Bronstein et~al.(2017)Bronstein, LeCun, Szlam, Vandergheynst, and
  Bruna]{Bronstein2017GeometricData}
Bronstein, M.~M., LeCun, Y., Szlam, A., Vandergheynst, P., and Bruna, J.
\newblock {Geometric Deep Learning: Going beyond Euclidean data}.
\newblock \emph{IEEE Signal Processing Magazine}, 34\penalty0 (4):\penalty0
  18--42, 11 2017.
\newblock ISSN 1053-5888.
\newblock \doi{10.1109/msp.2017.2693418}.
\newblock URL \url{http://arxiv.org/abs/1611.08097
  http://dx.doi.org/10.1109/MSP.2017.2693418}.

\bibitem[Cangea et~al.(2018)Cangea, Veli{\v{c}}kovi{\'{c}}, Jovanovi{\'{c}},
  Kipf, and Li{\`{o}}]{Cangea2018TowardsClassifiers}
Cangea, C., Veli{\v{c}}kovi{\'{c}}, P., Jovanovi{\'{c}}, N., Kipf, T., and
  Li{\`{o}}, P.
\newblock {Towards Sparse Hierarchical Graph Classifiers}.
\newblock 11 2018.
\newblock URL \url{http://arxiv.org/abs/1811.01287}.

\bibitem[Fey \& Lenssen(2019)Fey and Lenssen]{Fey/Lenssen/2019}
Fey, M. and Lenssen, J.~E.
\newblock Fast graph representation learning with {PyTorch Geometric}.
\newblock In \emph{ICLR Workshop on Representation Learning on Graphs and
  Manifolds}, 2019.

\bibitem[Gao \& Ji(2019)Gao and Ji]{gao2019graph}
Gao, H. and Ji, S.
\newblock Graph u-net, 2019.
\newblock URL \url{https://openreview.net/forum?id=HJePRoAct7}.

\bibitem[Gilmer et~al.(2017)Gilmer, Schoenholz, Riley, Vinyals, and
  Dahl]{gilmer2017neural}
Gilmer, J., Schoenholz, S.~S., Riley, P.~F., Vinyals, O., and Dahl, G.~E.
\newblock Neural message passing for quantum chemistry.
\newblock In \emph{Proceedings of the 34th International Conference on Machine
  Learning-Volume 70}, pp.\  1263--1272. JMLR. org, 2017.

\bibitem[Glorot \& Bengio(2010)Glorot and
  Bengio]{Glorot2010UnderstandingNetworks}
Glorot, X. and Bengio, Y.
\newblock {Understanding the difficulty of training deep feedforward neural
  networks}, 3 2010.
\newblock ISSN 1938-7228.
\newblock URL \url{http://proceedings.mlr.press/v9/glorot10a.html}.

\bibitem[Glorot et~al.(2011)Glorot, Bordes, and Bengio]{pmlr-v15-glorot11a}
Glorot, X., Bordes, A., and Bengio, Y.
\newblock Deep sparse rectifier neural networks.
\newblock In Gordon, G., Dunson, D., and Dudík, M. (eds.), \emph{Proceedings
  of the Fourteenth International Conference on Artificial Intelligence and
  Statistics}, volume~15 of \emph{Proceedings of Machine Learning Research},
  pp.\  315--323, Fort Lauderdale, FL, USA, 11--13 Apr 2011. PMLR.
\newblock URL \url{http://proceedings.mlr.press/v15/glorot11a.html}.

\bibitem[Hamilton et~al.(2017)Hamilton, Ying, and
  Leskovec]{hamilton2017inductive}
Hamilton, W., Ying, Z., and Leskovec, J.
\newblock Inductive representation learning on large graphs.
\newblock In \emph{Advances in Neural Information Processing Systems}, pp.\
  1024--1034, 2017.

\bibitem[He et~al.(2015)He, Zhang, Ren, and Sun]{He2015DelvingClassification}
He, K., Zhang, X., Ren, S., and Sun, J.
\newblock {Delving Deep into Rectifiers: Surpassing Human-Level Performance on
  ImageNet Classification}.
\newblock 2 2015.
\newblock URL \url{http://arxiv.org/abs/1502.01852}.

\bibitem[He et~al.(2016)He, Zhang, Ren, and Sun]{He_2016}
He, K., Zhang, X., Ren, S., and Sun, J.
\newblock Deep residual learning for image recognition.
\newblock \emph{2016 IEEE Conference on Computer Vision and Pattern Recognition
  (CVPR)}, Jun 2016.
\newblock \doi{10.1109/cvpr.2016.90}.
\newblock URL \url{http://dx.doi.org/10.1109/CVPR.2016.90}.

\bibitem[Kipf \& Welling(2016)Kipf and
  Welling]{Kipf2016Semi-SupervisedNetworks}
Kipf, T.~N. and Welling, M.
\newblock {Semi-Supervised Classification with Graph Convolutional Networks}.
\newblock 9 2016.
\newblock URL \url{http://arxiv.org/abs/1609.02907}.

\bibitem[Luzhnica et~al.(2019)Luzhnica, Day, and
  Lio']{Luzhnica2019CliqueClassification}
Luzhnica, E., Day, B., and Lio', P.
\newblock {Clique pooling for graph classification}.
\newblock 3 2019.
\newblock URL \url{http://arxiv.org/abs/1904.00374}.

\bibitem[Mishkin \& Matas(2015)Mishkin and Matas]{Mishkin2015AllInit}
Mishkin, D. and Matas, J.
\newblock {All you need is a good init}.
\newblock 11 2015.
\newblock URL \url{http://arxiv.org/abs/1511.06422}.

\bibitem[Niepert et~al.(2016)Niepert, Ahmed, and Kutzkov]{niepert2016learning}
Niepert, M., Ahmed, M., and Kutzkov, K.
\newblock Learning convolutional neural networks for graphs, 2016.

\bibitem[Simonovsky \& Komodakis(2017)Simonovsky and
  Komodakis]{Simonovsky_2017}
Simonovsky, M. and Komodakis, N.
\newblock Dynamic edge-conditioned filters in convolutional neural networks on
  graphs.
\newblock \emph{2017 IEEE Conference on Computer Vision and Pattern Recognition
  (CVPR)}, Jul 2017.
\newblock \doi{10.1109/cvpr.2017.11}.
\newblock URL \url{http://dx.doi.org/10.1109/CVPR.2017.11}.

\bibitem[Xu et~al.(2018)Xu, Li, Tian, Sonobe, Kawarabayashi, and
  Jegelka]{Xu2018RepresentationNetworks}
Xu, K., Li, C., Tian, Y., Sonobe, T., Kawarabayashi, K.-i., and Jegelka, S.
\newblock {Representation Learning on Graphs with Jumping Knowledge Networks}.
\newblock 6 2018.
\newblock URL \url{http://arxiv.org/abs/1806.03536}.

\bibitem[Ying et~al.(2018)Ying, You, Morris, Ren, Hamilton, and
  Leskovec]{Ying2018HierarchicalPooling}
Ying, R., You, J., Morris, C., Ren, X., Hamilton, W.~L., and Leskovec, J.
\newblock {Hierarchical Graph Representation Learning with Differentiable
  Pooling}.
\newblock 2018.
\newblock ISSN 18160948.
\newblock \doi{arXiv:1806.08804v3}.
\newblock URL \url{https://arxiv.org/pdf/1806.08804.pdf
  http://arxiv.org/abs/1806.08804}.

\bibitem[Zhang et~al.(2018)Zhang, Cui, Neumann, and Chen]{zhang2018end}
Zhang, M., Cui, Z., Neumann, M., and Chen, Y.
\newblock An end-to-end deep learning architecture for graph classification.
\newblock In \emph{Thirty-Second AAAI Conference on Artificial Intelligence},
  2018.

\end{thebibliography}
\bibliographystyle{icml2019}

\end{document}